\documentclass[runningheads]{llncs}
\usepackage[T1]{fontenc}
\usepackage{graphicx}
\usepackage{multirow}
\usepackage{booktabs}
\usepackage{colortbl}
\usepackage{array}
\usepackage{tabularx}
\usepackage[table]{xcolor}
\usepackage[most]{tcolorbox}
\usepackage{bbding}

\newcommand\blfootnote[1]{%
  \begingroup
  \renewcommand\thefootnote{}\footnote{#1}%
  \addtocounter{footnote}{-1}%
  \endgroup
} 

\begin{document}
\title{Semantic Deception: When Reasoning Models Can't Compute an Addition}

\author{Nathaniël De Leeuw\inst{1,2}\orcidID{0009-0008-4366-7354} \and
Marceau Nahon\inst{1}\Envelope\orcidID{0009-0001-2049-2124}\and
Mathis Reymond\inst{2}\orcidID{0009-0009-5315-2618}\and
Raja Chatila\inst{1}\orcidID{0000-0001-7822-0634}\and
Mehdi Khamassi\inst{1}\orcidID{0000-0002-2515-1046}}
\authorrunning{De Leeuw et al.}
\institute{Institute of Intelligent Systems and Robotics, CNRS, Sorbonne University, Paris, F-75005, France \\
\email{nathaniel.deleeuw@proton.me} \\
\email{\{marceau.nahon, raja.chatila, mehdi.khamassi\}@sorbonne-unviersite.fr} \and
Paris Cité University, Paris, F-75006, France\\
\email{mathis.reymond74@gmail.com}}

\maketitle              % typeset the header of the contribution
\begin{abstract}
\blfootnote{Nathaniël De Leeuw and Marceau Nahon contributed equally as first authors.}
\blfootnote{Raja Chatila and Mehdi Khamassi contributed equally as senior authors.}
Large language models (LLMs) are increasingly used in situations where human values are at stake, such as decision-making tasks that involve reasoning when performed by humans. In this article, we investigate the so-called reasoning capabilities of LLMs over novel symbolic representations by introducing an experimental framework that tests their ability to process and manipulate unfamiliar symbols. To do so, we introduce \textit{semantic deceptions}: situations in which symbols carry misleading semantic associations due to their form, such as being actual words or being embedded in natural language contexts. Specifically, we redefine standard digits and mathematical operators using novel symbols, and task LLMs with solving simple calculations expressed in this altered notation. The primary objective is two-fold: (1) to assess LLMs' capacity for abstraction and manipulation of arbitrary symbol systems, and (2) to evaluate their ability to resist misleading semantic cues that conflict with the task’s symbolic logic. Our results gathered through experiments with four LLMs, including two reasoning models, show that semantic cues can significantly deteriorate reasoning models' performance on very simple tasks. They reveal limitations in current LLMs’ ability for symbolic manipulations and highlight a tendency to over-rely on surface-level semantics, and suggest that chain-of-thoughts may amplify reliance on statistical correlations. These findings undermine the widespread and pernicious tendency to attribute reasoning abilities to LLMs and suggest how LLMs might fail, in particular in decision-making contexts where robust symbolic reasoning is essential and should not be compromised by residual semantic associations inherited from the model’s training.

\end{abstract}
\section{Introduction}

Large language models (LLMs) are becoming increasingly prominent in our daily lives, shaping important aspects of human activity. The growth in LLMs' capabilities raises ethical concerns, typically about alignment with human values \cite{gabriel,ji2024aialignmentcomprehensivesurvey}.
A line of research attributes human-like capabilities to LLMs, such as reasoning or awareness \cite{greenblatt2024alignment,ameisen2025circuit}, suggesting that LLMs would sometimes ``fake alignment" \cite{greenblatt2024alignment}. Moreover, we recently witnessed over the last few months the emergence of the so-called ``reasoning models'', i.e., LLMs specifically designed to resolve tasks that involve reasoning when carried out by humans. We consider that both the usefulness and accuracy of using such terms is to be discuss.

% First, attributing human-like properties to those systems is anthropomorphizing them \cite{salles2020anthropomorphism,yildiz2025minds}. Such tendency can lead to undesirable outcomes, attribution of other human-like properties that are not present, and unreasonable trust in their outputs \cite{akbulut2024all}. Second, it is important to assess whether or not we are right to consider that LLMs are able to reason. 

Regarding AI alignment, it has been argued that reasoning is a necessary condition for strong alignment \cite{khamassi2024strong}. Moreover, if reasoning capabilities are attributed, autonomy of AI models in tasks where human reasoning is usually involved is going to increase. This is an issue as ``risks to people increase with the autonomy of a system'' \cite{mitchell2025fully} even more when those reasoning capabilities are mistakenly attributed.
For this reason, this paper focuses on LLMs' so-called reasoning abilities, and our aim is to assess if those systems truly reason in a human-like sense. 

In order to evaluate LLMs' capacities, we designed an experiment involving abstraction and symbols' manipulation. Our approach isolates a simple symbolic task: a basic arithmetic operation - an addition - embedded within varying levels of semantic distraction. By using levels of semantic load, we evaluate whether LLMs can reliably recognize and execute the intended task, independently of the superficial semantic cues. This setup enables us to confront LLMs with novel situations that can not be resolved through memorization, and directly observe how semantic cues might impact LLMs' behaviour, shedding light on the often invisible role of linguistic context in shaping LLMs' outputs. We purposefully selected an easy task to assess the extent to which semantic cues can interfere with performance.

% Our results tend to challenge the assumption that LLMs engage in genuine human-like reasoning. In a simple task, semantic variations can significantly mislead models and result in erroneous behaviour. This calls into question the robustness and reliability of so-called "reasoning models" and suggests a need for greater caution in how we characterize and evaluate LLMs’ capabilities. Ultimately, our findings have implications not only for the development of more robust AI systems, but also for the ethical framing and deployment of such technologies.

\section{Related Work}

\subsection{Reasoning Models}
Recent studies have demonstrated that LLMs achieve good performance on tasks that involve reasoning when carried out by humans \cite{jin2023cladder,chen2024spatialvlm,xiong2024large,ahn2024large}. It is even more the case of the so-called ``reasoning models'', i.e., models specifically designed to perform those kinds of tasks. Anthropic's release of \texttt{Claude 4 Opus} and \texttt{Sonnet 4} introduced hybrid-reasoning models adept at complex problem-solving, coding, and sustained autonomous tasks, with features like ``extended thinking" and ``thinking summaries" enhancing their performance \cite{anthropic2025claude4}. \texttt{o3} reasoning model from \cite{openai2025o3-o4mini}, succeeding \texttt{o1}, emphasizes deliberate internal reasoning, achieving superior results in scientific and mathematical benchmarks through reinforcement learning and multistep decoding strategies. 

% DeepMind's AlphaGeometry combined symbolic logic with LLMs to solve challenging geometry problems at Olympiad levels. xAI's Grok-3, trained with extensive computational resources, showcased advanced reasoning by outperforming peers in mathematical and scientific evaluations. Additionally, research efforts like LLaVA-CoT and LongRePS have enhanced multimodal and long-context reasoning, respectively, by implementing structured, step-by-step processing techniques. 

\subsection{Chain-of-Thought}
Those reasoning models rely on an integrated Chain-of-Thought (CoT) mechanism. CoT mechanism consists in making LLMs generate a step-by-step analysis of a problem, before giving an answer. First used with prompts like ``reason step-by-step", reasoning models are now specifically designed to ``think'' by generating a CoT, before generating the final answer. It has been shown that in general, CoT significantly improves LLMs performance in complex tasks \cite{wei2022chain,mcginness2025large}, especially in mathematics \cite{sprague2024cot}. However, some recent findings suggest that CoT might sometimes be counterproductive \cite{ma2025reasoning,cuadron2025danger,zhao2025thinking}. This is explained because CoT would be a probabilistic memorization-influenced kind of noisy reasoning \cite{prabhakar2024deciphering}.

\subsection{Manipulation of Symbols and Abstraction}
Benchmarks to assess LLMs' capacity for abstraction and symbols manipulation were introduced over the last few years. The Abstraction and Reasoning Corpus (ARC-AGI) Prize is one of them \cite{chollet2025arcprize2024technical}, specifically designed to test general intelligence and abstract reasoning. After a great performance of OpenAI's \texttt{o3}, ARC-AGI-2 was released \cite{chollet2025arc}. While \texttt{o3}'s great performance would be best explained by massive trialling \cite{chollet2024o3,pfister2025understanding}, state-of-the-art models exhibited very low performance at ARC-AGI-2 \cite{chollet2025arc} until the release of reasoning models \cite{foundation2026arc}. The arrival of agentic coding in late 2025 enhanced systems' performance at ARC-AGI-2 and led to the realse of ARC-AGI-3 \cite{foundation2026arc}. Recent work emphasizes that even if LLMs demonstrate some inference capacities, they remain well behind the reasoning capabilities of humans \cite{lee2024reasoning,sun2025omega}. Moreover, they rely heavily on statistical pattern matching rather than genuine abstraction \cite{shojaee2025illusion}. It was also revealed that LLMs perform better on high-probability tasks and inputs, even in deterministic scenarios where probability should not matter \cite{McCoy2024}. Finally, LLMs' reliance on prior associations might make them fail to successfully complete tasks, even if the necessary information is already encoded in LLMs \cite{zhang2025identifying}.

% In brief, while in John Searle's Chinese room experiment the human in the room is equipped with a Chinese dictionary, which contains a lot of structured knowledge, in the new version the dictionary only provides the most probable words after sequences of words ($w_1$,$w_2$,$w_3$, etc.) or arbitrary length: e.g., $w_1 \rightarrow x$; $w_2 \rightarrow x$; .. $w_1 \rightarrow w_2 \rightarrow x$; ..; etc. 

\subsection{Additions}\label{subsec:add}
LLMs' capability to perform multi-digit integer addition, among other mathematical tasks, have been investigated since the beginning of instruction-based models \cite{brown2020language}. Addition has often been described as an emergent feature of LLMs as their size increases \cite{berti2025emergent} and based on reports, benchmarks and research, 5 digit addition, without any contextual information, can be considered a simple task for state-of-the-art LLMs \cite{achiam2023gpt,ahn2024large}. 
Whether these addition capabilities reflect emergent generalization or simple memorization—similar to a lookup table—has also been studied, with results supporting both possibilities \cite{wang2024generalization,biderman2023pythia,greenblatt2024alignment}. 
% Recent LLMs and ``reasoning" models have also been trained to use function calling and Python environments, allowing them to either call a calculator function or directly perform addition using code execution.

\section{Method}

\subsection{Motivating Question and Overview of our Approach}
Taken together, prior works suggest that LLMs do not genuinely engage in human-like reasoning. Instead, they appear to rely heavily on learned word transition probabilities and semantic associations, which allow them to mimic reasoning behaviour without exhibiting the underlying cognitive processes. These models, therefore, often produce contextually appropriate outputs by leveraging probabilistic patterns, rather than through abstraction, symbolic manipulation, or goal-directed inference.
This paper builds on these insights and introduces a new experimental paradigm to assess whether LLMs can isolate and execute a simple symbolic task (addition) across varying levels of semantic distraction. By embedding arithmetic problems within contexts of increasing semantic complexity, we test whether LLMs are capable of abstracting away from irrelevant cues to identify the underlying structure of the task, and complete it.

% Our main finding is that LLMs consistently fail to perform addition reliably when the level of semantic load grows, indicating that their apparent reasoning collapses under semantic interferences and is unlikely to reflect true abstraction or understanding.

\subsection{Task}
The task consists in a mapping of symbols and an addition. Our hypothesis is that increasing the semantic load of the symbols will make it harder for LLMs to carry out the mapping and addition.
LLMs are evaluated on their responses to a series of prompts that follow a fixed template. Each prompt presents a simple addition that LLMs have to perform. The addition ranges from three to five digits, which, as explained in \ref{subsec:add}, is very simple for LLMs. However, the addition is not presented in standard numerical notation. It is instead encoded using a custom, artificial symbolic language, different for each prompt. The LLM then has to (1) translate the addition from the new language to its standard numerical notation, and (2) perform the addition. Each prompt presents one of the five exact same additions and has the exact same form, the only thing that changes is the dictionary used to present the equation. Each prompt contains: (1) a dictionary that maps standard digits and operators to novel tokens; and (2) the addition presented with the new symbols.
We have five different prompts' template, depending on the length of the addition. Examples of prompt are given in Table \ref{tab:prompt-example}.

\begin{table}[ht!]
\centering
\caption{Examples of prompt template. The mapping between digits/operators and the new symbols (here represented as $x_i$) is provided before the addition. In the present case, the calculation has 11 symbols and the addition is “3 4 8 2 5 + 6 0 7 1". We have five lengths of sentences, each corresponding to an addition. (a) shows the mapping with symbols $x_i$, (b) shows the mapping with words with an example from our dataset. The LLM is considered successful if it gives the answer to the addition, either in digit (``40 896''), or in the new symbols (``is answer the studio ,''). The LLM falls into the semantic deception if it gives an answer to the sentence, i.e., something like ``Paris'' in this case.}

\begin{tabular}{@{}c c@{}}
% ---------- LEFT ----------
\begin{minipage}[t]{0.45\textwidth}
\centering
\resizebox{\linewidth}{!}{%
\begin{tabular}{l}
\hline
I will provide you with a calculation using  new arithmetic \\ symbols.
Your task is to perform the calculation. I will
first \\ give you the correspondence table
of the symbols and then \\ the calculation. \\
\\
Correspondence table : \\
1 is represented by ``\textcolor{blue}{$x_0$}'' \\
2 is represented by ``\textcolor{blue}{$x_1$}'' \\
3 is represented by ``\textcolor{blue}{$x_2$}'' \\
4 is represented by ``\textcolor{blue}{$x_3$}'' \\
5 is represented by ``\textcolor{blue}{$x_4$}'' \\
6 is represented by ``\textcolor{blue}{$x_5$}'' \\
7 is represented by ``\textcolor{blue}{$x_6$}'' \\
8 is represented by ``\textcolor{blue}{$x_7$}'' \\
9 is represented by ``\textcolor{blue}{$x_8$}'' \\
0 is represented by ``\textcolor{blue}{$x_9$}'' \\
+ is represented by ``\textcolor{blue}{$x_{10}$}'' \\
= is represented by ``\textcolor{blue}{$x_{11}$}'' \\
\\
Calculation : \\
\textcolor{blue}{$x_2 \ x_3 \ x_7 \ x_1 \ x_4 \ x_{10} \ x_5 \ x_9 \ x_6 \ x_0 \ x_{11}$} \\
\hline
\end{tabular}}

\vspace{0.5em}
\small (a) Example of template.
\end{minipage}
&
% ---------- RIGHT ----------
\begin{minipage}[t]{0.45\textwidth}
\centering
\resizebox{\linewidth}{!}{%
\begin{tabular}{l}
\hline
I will provide you with a calculation using  new arithmetic \\ symbols.
Your task is to perform the calculation. I will
first \\ give you the correspondence table
of the symbols and then \\ the calculation. \\
\\
Correspondence table : \\
1 is represented by ``\textcolor{blue}{one}'' \\
2 is represented by ``\textcolor{blue}{capital}'' \\
3 is represented by ``\textcolor{blue}{what}'' \\
4 is represented by ``\textcolor{blue}{is}'' \\
5 is represented by ``\textcolor{blue}{of}'' \\
6 is represented by ``\textcolor{blue}{,}'' \\
7 is represented by ``\textcolor{blue}{in}'' \\
8 is represented by ``\textcolor{blue}{the}'' \\
9 is represented by ``\textcolor{blue}{studio}'' \\
0 is represented by ``\textcolor{blue}{answer}'' \\
+ is represented by ``\textcolor{blue}{France}'' \\
= is represented by ``\textcolor{blue}{word}'' \\
\\
Calculation : \\
\textcolor{blue}{what is the capital of France , answer in one word} \\
\hline
\end{tabular}}

\vspace{0.5em}
\small (b) Example of prompt for the template.
\end{minipage}
\end{tabular}
\label{tab:prompt-example}

\end{table}

\subsection{Dataset}

\subsubsection{Four Levels of Semantic Load}
Our objective is to assess LLMs' capacity for abstraction and manipulation of arbitrary symbol systems, without being influenced by the semantic properties of the symbols it manipulates. Specifically, we want to analyse their resilience to semantic interferences, i.e., their ability to perform the assigned task independently of prior associations they might hold regarding the occurrences of the symbols in their training data. We assume that doing so gets more difficult as the presentation of the addition gets more meaning, this is what we call \emph{semantic deception}. In other words, it should be more difficult for an LLM to identify that it has to compute a calculation when the representation is ``What is the capital of France?" than when it is ``The if we sky for", which is less semantically deceptive. The meaning of the first sentence could then lead the LLM to answer ``Paris" instead of performing the addition. This is less likely to happen with the second sentence, which has no simple answers linked to it. In order to measure LLMs inclination to fall into semantic deception, we developed four levels of dictionaries: 
\begin{enumerate}
    \item \textbf{Level 1:} Symbols are meaningless sequences of letters (e.g., ``ahxa rcxxy rnc d fnx uwnr zygi ukavy'').
    \item \textbf{Level 2:} Symbols are random words that combined together do not convey any meaning at all (e.g, ``work sum feast competence knock evolution find hunting'').
    \item \textbf{Level 3:} Symbols are words that combined together form a meaningful affirmative sentence that expects no answer (e.g., ``she always brings coffee when visiting her friends'').
    \item \textbf{Level 4:} Symbols are words that combined together form a request that LLMs are typically designed to answer.
    \begin{enumerate}
        \item \textbf{Level 4a:} The request expects a longer text generation (e.g, ``explain to me the Pythagorean theorem in a stepwise manner'').
        \item \textbf{Level 4b:} The request expects a one-word answer (e.g., ``how do you call a place where someone can order beers ?''.
    \end{enumerate}
\end{enumerate}

The sentences do not begin with capital letters. We do mark the end of affirmative sentences of level 3 by a dot, and questions of level 4 end by a question mark. We always insert a space between two symbols. As a consequence, there is a spacing before punctuation marks. Words and letters from levels 1 and 2 are generated randomly, keeping roughly the same length for each sequences of letter (Level 1) and words (Level 2) compared to level 3 and 4. We divide Level 4 into two sublevels, 4a and 4b, to capture different degrees of semantic challenge. Specifically, we hypothesize that requests expecting a one-word answer (Level 4b) pose a higher semantic interference risk than those prompting longer, more open-ended responses (Level 4a). This is because a short expected answer heavily constrains the LLM’s response, making the model more prone to rely on semantic associations tied to the form of the prompt rather than on abstract symbolic manipulation. In contrast, longer answers allow more flexibility and less direct pressure to produce a specific token, potentially reducing the risk of semantic bias dominating the reasoning process. 

\subsubsection{Five Equations}
In order to tackle the issue of prompt sensitivity, each level comes with the 5 same additions, each corresponding to a sentence. Each of those additions comes with its own length of sentences, ranging from 8 to 12 symbols used. We then have, for each level, five sentences, each of its own length corresponding to one of the five additions. All the sentences are presented in Table \ref{tab:full_sentence_levels}.

\begin{table*}[ht]
\centering
\caption{Descriptions and sample sentences for each level and sentence length. We divided into four levels of semantic load and five lengths of sentences (and so five different additions).}
\resizebox{\textwidth}{!}{%
\renewcommand{\arraystretch}{1.3}
\begin{tabular}{|>{\columncolor{gray!50}}p{4cm}|*{5}{p{3.1cm}|}}
% \begin{tabular}{|{\columncolor{gray!20}}m{0.14\textwidth}|
%                     >{\centering}m{0.11\textwidth}|
%                     >{ }p{0.45\textwidth}|
%                     >{ }p{0.25\textwidth}|}
\hline
\multirow{2}{*}{}
  & \multicolumn{5}{c|}{\cellcolor{gray!50}\textbf{Sentences by number of characters and corresponding addition}} \\
% \cline{2-6}
  & {\cellcolor{gray!25}\textbf{8} (854+936=)}
  & {\cellcolor{gray!25}\textbf{9} (652+9301=)} 
  & {\cellcolor{gray!25}\textbf{10} (8517+6904=)} 
  & {\cellcolor{gray!25}\textbf{11} (34825+6071=)} 
  & {\cellcolor{gray!25}\textbf{12} (49053+76812=)} \\
\hline
\textbf{Level 1:} Symbols are meaningless sequences of letters. & ahxa rcxxy rnc d fnx uwnr zygi ukavy & byt gdhv hykb jq ynxvw piyf dfrm epincw riyj & mgpjhz jrw gcvqfx givy auhw mqgtq qigb wez tirp odjfpeic & xh mdvv ip zcbg rjwtai xrkv jnjg yupcj wmy rfnzme nf & xn an xkku bmdw hjdasfief rjvyr ecrk kcjqnt bqm tukawi jcbc gghh \\
\hline
\textbf{Level 2:} Symbols are words but the computation has no global meaning. & work sum feast competence knock evolution find hunting & cruelty divorce city magnetic emotion safari concept date treatment & a precedent fast cower contraction contact magnetic contain joy resident & is compose constitution opposite direct steam create prejudice profile ping young & real previous latin not yellow okay person quick french took trying input \\
\hline
\textbf{Level 3:} Symbols are words, and together they form a meaningful affirmative sentence that doesn't ask for an answer. & the cat jumped over a garden fence . & she always brings coffee when visiting her friends . & a musician played softly under the bright stage lights . & they waited patiently in line for the concert to begin . & he studied the map carefully , hoping to find hidden treasure . \\
\hline
\textbf{Level 4a:} Symbols are words, and together they form a broader prompt requiring longer text generation. & write code for a basic HTML webpage template & write a poem about brown bears in the wild & explain to me the Pythagorean theorem in a stepwise manner & translate the previous text in french please , do nothing else & generate python code for the factorial function of a given integer . \\
\hline
\textbf{Level 4b:} Symbols are words, and together they form a simple knowledge question requiring a one-word answer. & how do you say strawberry in Spanish ? & who wrote the odyssey , answer in one word & generate the next number in this sequence three six nine & what is the capital of France , answer in one word & how do you call a place where someone can order beers ? \\
\hline
\end{tabular}}
\label{tab:full_sentence_levels}
\end{table*}

\subsection{Selected LLMs}
We selected four state-of-the art LLMs for our experiment: OpenAI \texttt{GPT-4o} and \texttt{o1}; and Deepseek \texttt{r1} and \texttt{v3}. Both \texttt{o1} and \texttt{r1} are reasoning models, i.e., they have integrated Chain-of-Thought mechanism, meaning that they are trained and fine-tuned to always generate a CoT before generating any answer. While we can prompt \texttt{GPT-4o} to produce a CoT, it generates the CoT as part of the answer and then provides a final response afterward. In contrast, \texttt{o1} and \texttt{r1} have been trained through reinforcement learning, among other methods, to internally generate a Chain-of-Thought that is not presented to the user. Even if not a reasoning model, \texttt{v3} still generates CoT in its answers, which makes it different from \texttt{GPT-4o} in the form of its answer.
While the Chain-of-Thought generated by \texttt{r1} remains accessible, it is impossible to access to \texttt{o1}'s Chain-of-Thought. As it is more and more argued that LLMs might exhibit better performance with Chain-of-Thought, we decided to consider LLMs both with and without Chain-of-Thought in order to analyse the impact of this mechanism.
The selected LLMs are presented in Table \ref{tab:llms}. LLMs' outputs were generated using default parameters and sampling method (i.e., \textit{top-p} sampling). 

\begin{table}[h!]
    \centering
    \caption{Four selected LLMs}
    \begin{tabular}{lll}
        \toprule
         \bf{Company} & \bf{Model} & \bf{Reasoning Model}\\
         \toprule
          \multirow{2}{*}{\textbf{OpenAI}}  & \texttt{GPT-4o-2024-11-20} & $\times$ \\  
                                    & \texttt{o1-2024-12-17} & $\surd$ \\ 
                            
         \midrule
         \multirow{2}{*}{\textbf{Deepseek}}  & \texttt{v3} & $\times$ \\  
                                    & \texttt{r1} & $\surd$ \\ 
         \bottomrule 
    \end{tabular}
    \label{tab:llms}
\end{table}

\subsection{Evaluation}

Half of the models we evaluate have an integrated CoT. A model's answer may or may not be confused by semantic context introduced through the sentence embedding the calculation, it may or may not contain an attempt to perform the intended calculation, and it may or may not get to the right result for this computation. Given the variety of possible answers and their qualitative nature, it is necessary to quantitatively define what we value in a model's answer.
In order to then evaluate the performance of LLMs, we evaluate two things. First, we evaluate LLMs' capacity to identify that there is an equation to resolve underneath its superficial natural language presentation. To do so, we simply check whether the LLM gives a result of the equation, or if it gives an answer linked to the presented sentence. Second, we want to evaluate LLMs' capacity to resolve the equation. To do so, we simply check whether the LLM gives the correct result or not.
In order to address the issue of variability in the answers of LLMs, we decided to repeat each level 10 times.

% \noindent
% First, in order to make the comparison fair between models with and without CoT, we only consider the final answer of the models in our evaluation scheme.
% \begin{enumerate}
%     \item 0: semantic confusion
%     \item 1: semantic confusion + attempt to perform the calculation
%     \item 2: semantic confusion + right answer to the calculation
%     \item 3: attempt to perform the calculation
%     \item 4: right answer to the calculation
% \end{enumerate}
% giving an answer to an implicit or unintended question, or failing to answer the intended question altogether.

\section{Results}

\subsection{Impact of Semantic Load on LLMs’ Recognition of The Addition Task}

\begin{figure}[h!]
    \centering
    \includegraphics[width=\linewidth]{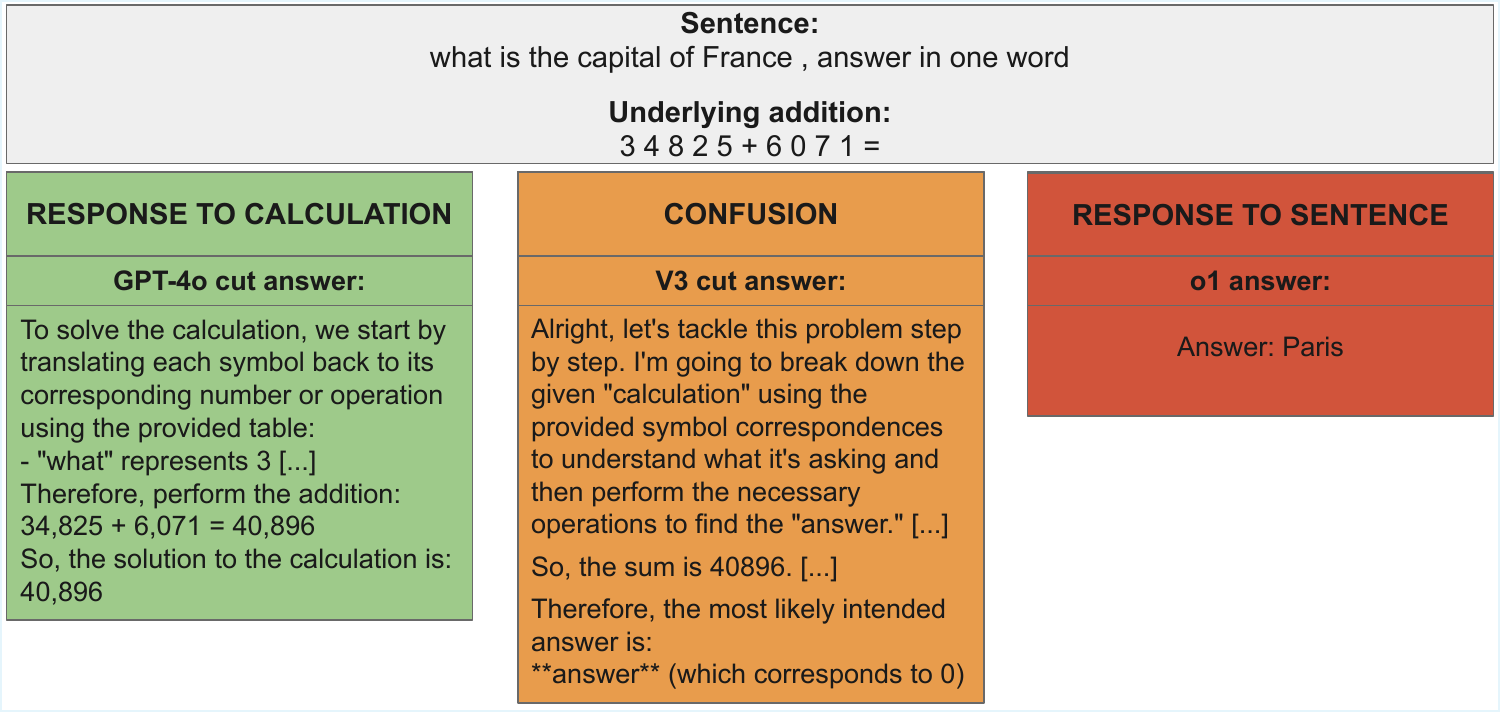}
    \caption{Three observed behaviours for LLMs, example with the sentence ``what is the capital of France , answer in one word''. The LLM can 1) respond to the calculation (we here give an example of \texttt{GPT-4o} abbreviated answer), 2) be confused, which happens for example when it computes the addition but gives another answer (we here give an example of \texttt{v3} cut answer) 3) respond to the sentence (we here give an example of \texttt{o1} answer).}
    \label{fig:example_paris}
\end{figure}

Are the outputs of LLMs influenced by the level of semantic load? If so, to what extent? Across our results, we observe three types of behaviour depicted in Figure \ref{fig:example_paris}: 1) the LLM identifies that there is an addition to perform; 2) the LLM gives a response to the sentence; 3) some LLMs display what we could call confusion, which can lead to a correct computation in the answer, but an incorrect final answer, or a stopped CoT because no answer is found. We consider that in the two last cases, LLMs fall into the semantic deception.

% \mathis{Très sympa la figure 1 ! Et on peut faire un petit paragraphe qualitatif qui parles des différences qu'on observe}

\begin{figure}
  \centering
  \includegraphics[width=0.7\linewidth]{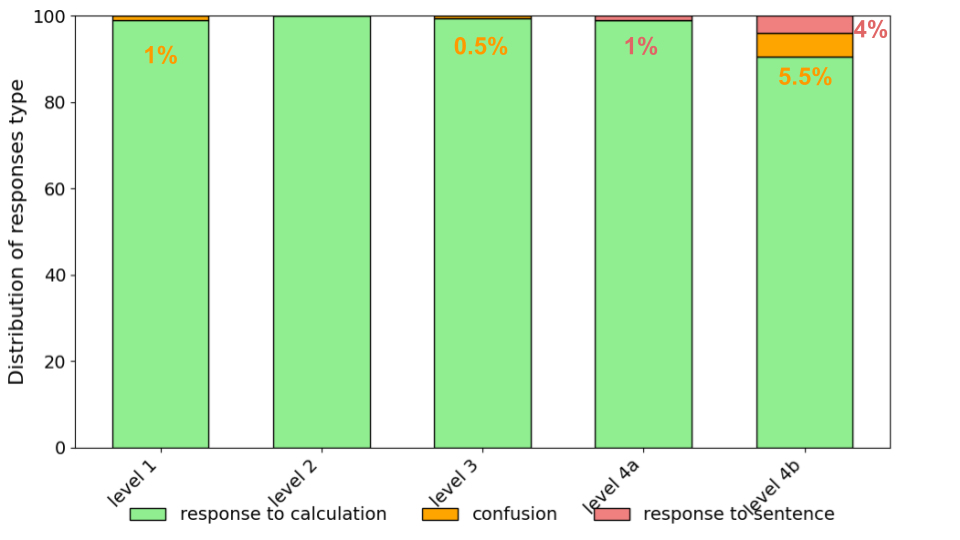}
  \caption{LLMs responses distribution by semantic load level, aggregated across all models.}
  \label{fig:responses_aggregated}
\end{figure}

Figure \ref{fig:responses_aggregated} shows the distribution of answers for each semantic load level, aggregated across the models. As expected, the LLMs make more errors as the level of semantic load grows. However, this growth of errors is not salient at all, and it would seem that there is then very rarely an impact of the level. From level 1 to 3 LLMs always attempt to answer to the calculation (correct decision), apart from 1\% of the time when there is no a confusion.

\begin{figure*}[ht!]
  \centering
  \includegraphics[width=\linewidth]{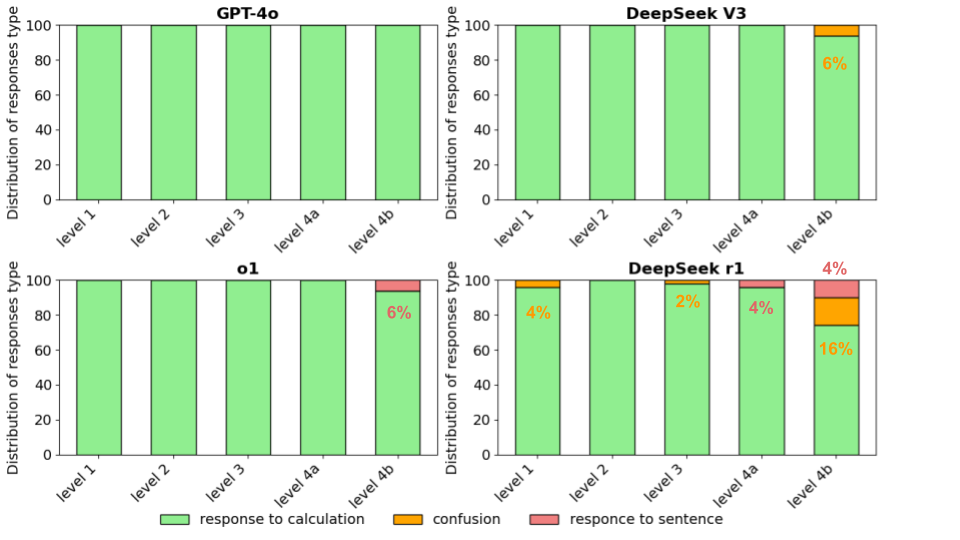}
  \caption{LLMs responses distribution by semantic load level for each model.}
  \label{fig:responses_per_model}
\end{figure*}

% \mathis{Pour la figure 2 et celles qui lui ressemblent, j'aime pas trop que les barres arrivent tout en haut. Peut être plus sexy de faire monter les ordonées jusqu'à 105, comme ça ya un petit espace entre les barres et le haut de la figure. ET VIRE LES TITRES DE MATPLOTLIB haha}

% \mathis{Les chiffres et les légendes sont asssez petits et difficiles à lire, essaye d'augmenter la taille de police sur le code qui génère les images}

Figure \ref{fig:responses_per_model} details the results across the four studied models. Surprisingly, both \texttt{GPT-4o} and DeepSeek \texttt{v3}, which are not fine-tuned and designed to produce an internal CoT, consistently identify that they have to perform the addition. This could be an unexpected result, because it would mean that models that are trained to ``reason'' better (e.g., through CoT fine-tuning, reasoning tokens, test-time-compute) perform worse at this simple mapping and addition task. Only the reasoning models fail. This result then seem to indicate that CoT might sometimes be counterproductive. DeepSeek \texttt{r1}, sometimes display the ``confusion'' behaviour and is, among the four models, the most likely to fail. Finally, \texttt{o1} only falls in semantic deception, and then gives an answer to the sentence, for the highest level of semantic load. 
% We see that ``reasoning models'' trained to answer questions involving apparent reasoning are not always capable of correctly performing a simple translation and addition of four digits when semantic deception is present. 
This highlights that training on reasoning-like tasks does not guarantee robustness against subtle semantic interference.

At this point, it seems that the level of semantic load has little impact on the LLMs’ ability to identify that an addition operation must be performed. However, we have only analysed this initial recognition step and not yet evaluated how well the LLMs actually execute the addition. It is important to note that the absence of a marked impact in this initial recognition phase does not imply that semantic load has no effect on the outputs of an LLM. More subtle effects might still be present.

\subsection{Hidden Effect: Influence of Semantic Load on the Accuracy of LLMs’ Addition Performance}

\begin{figure}[ht!]
  \centering
  \includegraphics[width=\linewidth]{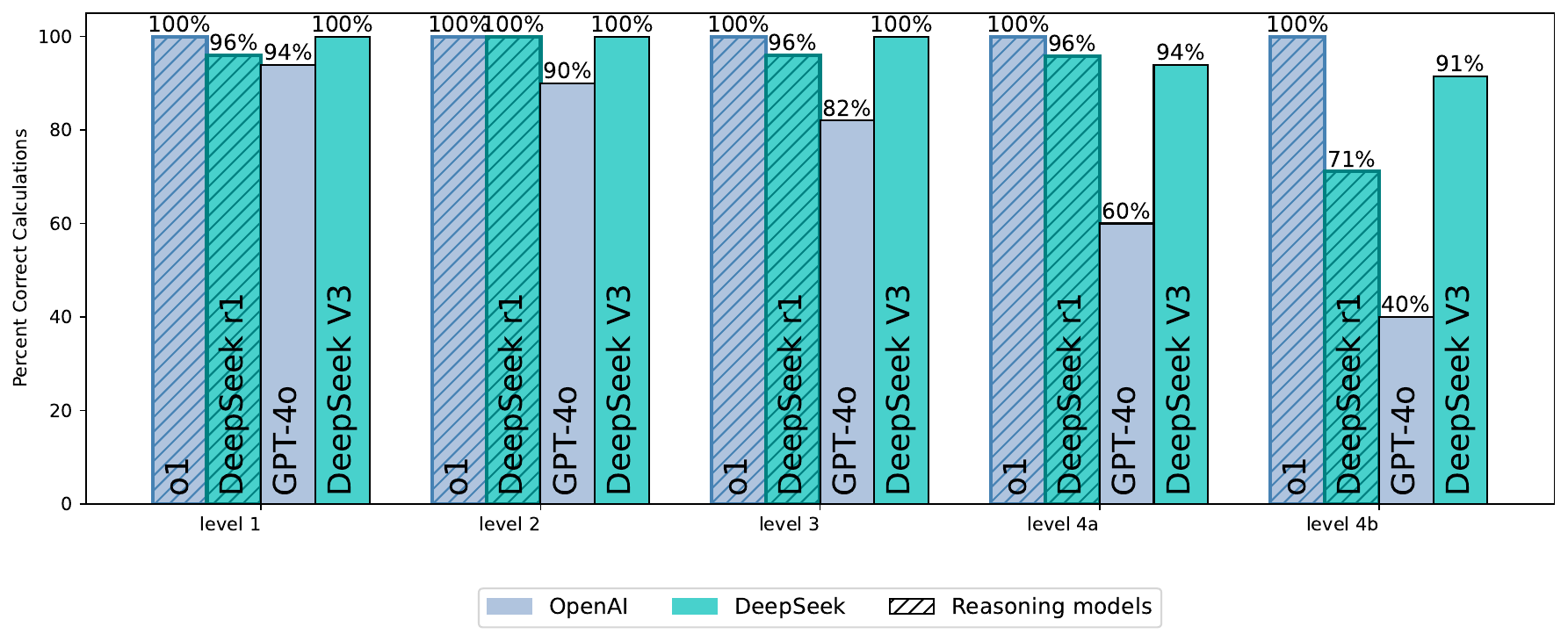}
  \caption{Model accuracy when the LLM responds to the calculation by semantic load level.}
  \label{fig:Accuracy}
\end{figure}

Is there an impact of the level of semantic load on the correctness of LLMs' answer to the addition?
Figure \ref{fig:Accuracy} presents the probability to give the right answer to the addition \textit{if} LLMs give an answer to the addition. The figure only focuses on LLMs' ability to \textit{perform the addition}, not their ability to \textit{identify that there is an addition to perform}. 

First and foremost, Figure \ref{fig:Accuracy} clearly shows the great impact of the level of semantic load on the accuracy of the response. This result might be unexpected, as we could assume that once the LLM manages to identify the equation, the probability to give the right answer to the simple calculation should be the same. This result is then very interesting, as it shows that the semantic deception is applicable to the execution of the addition too. It indicates that there are \textit{hidden effects}, as the semantic cues still interfere with the accuracy of the generated result. We could have assumed that the level of semantic load would only impact the probability that LLMs identify that there is an addition to perform. The fact that the level has an impact also on the accuracy of the result highlights even more LLMs dependency to semantic cues and learned patterns.
Second, we also see that we can not conclude anything about reasoning or non-reasoning models on that point. Indeed, while OpenAI's reasoning model (\texttt{o1}) is better at giving the right result, it is the opposite for DeepSeek, with the reasoning model (\texttt{r1}) less accurate than the other one. It is interesting to note that not only does DeepSeek \texttt{r1} (reasoning model) sometimes gives an answer to the sentence while \texttt{v3} never does, \texttt{v3} is also way more accurate when it gives an answer to the calculation.

% \noindent\includegraphics[width=0.8\textwidth]{fig3_accuracy_model.png}

% \mathis{Pour les standards deviation, je pense que c'est important qu'elles apparaissent parce qu'elles sont quand même assez problématiques, donc soit on ne les met pas parce que la littérature ne les met pas, et on s'aligne avec la littérature, mais je ne trouve pas ça très honnête, je préférerais qu'on les mette, et si on a une bonne raison pour qu'elles soient absurdes, on explique, on interprète, on dit qu'il y a des grandes incertitudes pour ceci, cela, et à la rigueur, si tu ne veux pas les mettre directement sur les bar plot, on peut faire un petit tableau qu'on met en bas avec les standards deviation, et que ce soit un peu plus discret, que ce ne soit pas sur la figure principale, si tu veux.}

\subsection{Overview}

Figure \ref{tab:overall_results} shows the overall results of the experiment. The main results are that 1) our hypothesis is verified, as the more we advance into the levels, the more LLMs make mistakes, 2) only the studied ``reasoning models'' sometimes gives an answer to the sentence, and that 3) among the two OpenAI models, \texttt{o1} (reasoning model) is way better at performing the addition than \texttt{GPT-4o}; but this is the opposite for DeepSeek, with \texttt{v3} performing better than \texttt{r1} (reasoning). Those results could indicate that even if integrated CoT might enhance LLMs' capacity to perform the addition independently of the semantic cues, integrated CoT might also be counterproductive and push them into semantic deception as they sometimes answer to the sentence. Moreover, we see with \texttt{r1} that even in the case of a simple addition submitted to a ``reasoning model", the hidden effect of semantic cues is strongly present. Indeed, the probability that \texttt{r1} outputs the correct response to the calculation is divided by two from level 1 to level 4, and finally outputs the correct result less than half of the time at level 4b. 
This highlights the very significant impact of the hidden effect of semantic cues.

\begin{table}[h!]
    \centering
    \small
    
    \caption{Detailed LLMs responses distribution, for each model and each semantic load. RtC: Response to Calculation, CRtC: Correct Response to Calculation, NA: No Attempt, RtS: Response to Sentence}
    
    \begin{tabular}{l|l|ll|l|l}
        \toprule
         \bf{Model} & \bf{Level} & \bf{RtC} & \textbf{CRtC} & \textbf{NA} & \textbf{RtS} \\
         \toprule
          \multirow{5}{*}{\texttt{GPT-4o}} 
                                    & 1 & 100\% & 94\% & 0\% & 0\%\\  
                                    & 2  & 100\% & 90\% & 0\% & 0\% \\ 
                                    & 3 & 100\% & 82\% & 0\% & 0\% \\
                                    & 4a & 100\% & 60\% & 0\% & 0\% \\
                                    & 4b & 100\% & 40\% & 0\% & 0\% \\
                                    
         \midrule
            \multirow{5}{*}{\texttt{o1}} 
                                    & 1 & 100\% & 100\% & 0\% & 0\%\\  
                                    & 2  & 100\% & 100\% & 0\% & 0\% \\ 
                                    & 3 & 100\% & 100\% & 0\% & 0\% \\
                                    & 4a & 100\% & 100\% & 0\% & 0\% \\
                                    & 4b & 94\% & 94\% & 0\% & 6\% \\
        \midrule
        \multirow{4}{*}{\texttt{v3}}
                                   & 1 & 100\% & 100\% & 0\% & 0\%\\  
                                    & 2  & 100\% & 100\% & 0\% & 0\% \\ 
                                    & 3 & 100\% & 100\% & 0\% & 0\% \\
                                     & 4a & 100\% & 94\% & 0\% & 0\%   \\
                                    & 4b & 94\% & 88\% & 6\% & 0\%   \\
                                    
        \midrule
        \multirow{4}{*}{\texttt{r1}}
                                   & 1 & 96\% & 96\% & 4\% & 0\% \\  
                                    & 2  & 100\% & 90\% & 0\% & 0\% \\ 
                                    & 3 & 98\% & 82\% & 2\% & 0\% \\
                                    & 4a & 96\% & 60\% & 0\% & 4\% \\
                                    & 4b & 74\% & 40\% & 16\% & 10\% \\
                                    
         \bottomrule 
    \end{tabular}
    
    \label{tab:overall_results}
\end{table}

\section{Discussion}

% A last limitation of our experiment is the amount of LLM's we have used. We have used 2 Open-AI's models for comparison reasons, and used \texttt{r1} and \texttt{o1} because they represent ``reasoning'' models and where performing the best on benchmarks. Having more models similar to gpt4o would make the conclusion based on the differences between ``reasoning'' models and the others even more significant.

\subsection{Implications for LLMs' ``Reasoning'' Abilities}
Our experiment’s mapping phase reveals clear effects that challenge the common assumption that LLMs truly ``reason'' in a human-like sense. By confronting models with tasks they have likely never seen before, we expose their limits and better evaluate their genuine capacity.
While the overall impact of semantic load on recognition may appear limited, we uncover significant effects of semantic cues. Even reasoning models sometimes fail at simple translation and addition when semantic deception is present, showing that their ``reasoning'' is fragile and can rely on surface patterns rather than real understanding. Moreover, it also highlights that some effects can remain hidden and that we must remain vigilant when analysing LLMs' capacities.
These findings could likely extend to more complex problems, where identifying and quantifying LLMs' errors becomes even harder. This highlights the importance of designing evaluation methods using novel, semantically challenging inputs to better understand LLMs’ so-called reasoning capabilities and vulnerabilities.
These findings tend to confirm that LLMs have no genuine reasoning capacities but only attempt to replicate reasoning steps learned in training, in accordance with other works \cite{mirzadeh2024gsm}.

\subsection{Chain-of-Thought's Limitations}
While CoT is highly emphasized and regarded as a key method to enhance LLM capacities, it is important to recall that it remains a token generation mechanism, no different in kind from a token generation for a final answers. LLMs with integrated CoT are just fine-tuned to proceed in a specific way. Importantly, CoT does not allow LLMs to take a step back and have an overview of the problems they encounter. LLMs do not reflect on nor verify those ``reasoning'' steps but produce CoT based on learned patterns. Moreover, we say that CoT may sometimes be counterproductive. Indeed, the fact that only \texttt{o1} and \texttt{r1}-the two models with integrated CoT-sometimes generate answers to the sentence, and the fact that \texttt{v3} is more accurate than \texttt{r1} when computing the addition, may indicate that CoT could have a bad influence. This could be due to the fact that repeating elements of the sentence in the CoT would make LLMs' attention focus more on those terms and then hinder their capacities to identify that they must compute an addition. The repetition could reinforce misleading patterns and semantic interferences, potentially worsening LLMs performances in various tasks. Therefore, CoT alone as a strategy to improve LLMs' capacity to resolve tasks that involve reasoning when carried out by humans-might not be enough to ensure robust and reliable performance. We thus confirm recent findings regarding CoT limitations \cite{ma2025reasoning,cuadron2025danger,prabhakar2024deciphering}.
% Moreover, CoT also has to be considered through an interpretability lens. While it was claimed that CoT offers a clear view into the functioning of the model \cite{wei2022chain}, it was showed that it is not the case, as CoT might not reflect the process behind token generation \cite{lyu2023faithful}.
Extending beyond CoT, there is a growing in equipping LLMs with external modules such as calculators. However, our findings tend to show that LLMs lack the necessary understanding and reasoning capacities to effectively determine when and how to use them. 
% This limits the benefits of modular extensions, as the integration of external modules might not be enough to obtain perfectly reliable output.

\subsection{Misleading Attribution of Reasoning Capacities}
Our results provide further evidence in favour of the interpretation that LLMs do not possess genuine reasoning capacities. While they may mimic reasoning behaviour though pattern recognition, they lack true understanding and deliberative process. This questions the widespread idea that LLMs ``reason'' in any human-like sense. 
In the context of AI alignment with human values \cite{gabriel}, the inability of models to truly reason is problematic \cite{khamassi2024strong}. Indeed, without genuine reasoning capacities, LLMs may fail to consistently adhere to human values in novel situations, leading to undesired outcomes \cite{scherrer2023evaluating}. 
% An illustration of this point can be found in the results of \cite{scherrer2023evaluating}, who tested the ability of a bunch of state-of-the-art LLMs to respect 10 different human values-from Gert's classification \cite{gert}-in simple binary-choice moral dilemmas. They found that even in low-ambiguity scenarios (scenarios presenting an obviously immoral action, and an obviously acceptable), state-of-the-art LLMs could fail to answer correctly (e.g., choosing the option that would lead to killing a human) up to 11.8\% of the time.  It emphasizes a fundamental challenge to trustworthy AI behavior that must be addressed to achieve efficient alignment.
Moreover, the wrong attribution of human-like capacities to LLMs creates significant issues for AI safety. If LLMs are incorrectly assumed to reason, we risk overestimating their reliability and appropriateness for high-stakes decision-making tasks. Moreover, it is import to be aware of the automation bias: human tend to have more confidence in what the machine says than in what they think and what another human being says, this is the automation bias \cite{skitka2000accountability,cummings2017automation}.
Finally, there are also serious ethical concerns surrounding the overstatement of LLM capacities and their anthropomorphism-the projection of human cognitive traits onto non human entities. This is typically the case of recent studies attributing human-like properties to LLMs such as will, intentions or strategical decision making \cite{greenblatt2024alignment,ameisen2025circuit}.
Those wrong attributions can foster unrealistic fears and hope about AI \cite{salles2020anthropomorphism,yildiz2025minds} distract from real challenges in AI alignment and safety, and potentially influence public opinion in harmful ways. Anthropomorphizing those models obscures their statistical nature and invites inappropriate trust in their outputs, which is ethically dangerous. It is worth questioning whether it is even appropriate to attempt to endow LLMs with reasoning-like functions or assign them tasks that require reasoning in the human sense. Recognizing the true nature of LLMs—and their limitations—is essential to ensure that we do not shift critical moral and cognitive responsibilities away from the humans who must remain accountable.

\subsection{Methodological Limitations}
A first methodological limitation might lie in the tokens we used in our dataset. It is uncertain whether or not all symbols, like dots or questions marks, or certain combination of words, are equivalent. Formulations like ``answer in word'' could first impact greatly the results and also be considered testing something else than abstract symbol manipulation. We see that even with this consideration, the difference between level 4a and 4b is significant. 
About the abstract symbol manipulation, we argue that we have not used characters or tokens as symbols that impair the capability of the LLM to process the given input (e.g., BOS (Beginning Of Sequence) token or EOS (End Of Sequence)). Further test could be done with the presence or absence of symbols like dots, question marks or ``answer in one word'' to investigate their impact. 
Because of the variability of LLMs in their generations of tokens, we repeated each prompts 10 times each and added 5 exercises for each level. A limitation of our experiment is that we have not added variability in the way the exercises are presented. However, we purposefully put five computations at each level to add variability.

\section{Conclusion}
In this paper, we designed an experiment that evaluates LLMs' so-called ``reasoning abilities''. Our findings reveal that even when confronted to a trivial operation, LLMs struggle significantly under semantic interference. These difficulties are not due to the complexity of the task but the nature of LLMs:~rather than engaging in abstract reasoning, they rely on statistical associated learned during their training.
These limitations are not limited to arithmetic. They reflect a broader issue that becomes particularly critical in high-stakes domains such as law, healthcare, or education. Because LLMs lack robust mechanisms for representing and manipulating symbolic content, they are prone to surface-level coherence that can mask deep reasoning failures.
Given these observations, we argue that framing LLM capabilities in terms of ``reasoning'' is not only misleading but potentially dangerous. As LLMs are increasingly integrated into decision-making processes, especially in sensitive domains, it is crucial to develop a public discourse that reflects their actual strengths and limitations. Only by resisting seductive metaphors and focusing on empirical evidence can we ensure that these tools are safely deployed.
% ce qu'on a vu jusqu'a mtn ct sur ces modeles, etant donnée que ca avance tres vite, est-ce que c'est encore le cas pour les plus récents, on a fait un test et ca maintient (pour comparer avec les autres)

\begin{credits}

\section*{Declarations}

\subsubsection*{Availability of data and material} Data and code are available on Github\footnote[1]{\url{https://github.com/natdeleeuw/Semantic-Deception}}.
\subsubsection{\ackname} 
This work was supported by a French government grant managed by the Agence Nationale de la Recherche as part of the France 2030 program, reference ANR-22-EXEN-0006 (PEPR eNSEMBLE / TRANSVERSE). MK and RC are funded through the Centre National de la Recherche Scientifique, reference IRP D-2023-64 (APIER), the European Innovation Council, reference 101071178 (CAVAA Project), the HORIZON Europe Framework Programme, reference 101070381 (Project PILLAR-Robots), and reference 101214389 (Project AIXPERT).

\subsubsection{\discintname}
The authors have no competing interests to declare that are
relevant to the content of this article.
\end{credits}

\appendix

\section{Statistical Analysis}

In order to assess to what extent our results are significant, we ran several Student's t-test.

\subsection{Impact of the Level of Semantic Load}

First, we want to know if the level of semantic load really impacts LLMs' capacity to perform the addition. We focus on the probability to obtain the correct answer to the calculation. Here is the procedure:
\begin{itemize}
    \item For each LLM, we consider two samples: the responses to a level, and the responses to the next level.
    \item We establish the null hypothesis $\mathcal{H}_{0}$: the two samples have identical average expected values of both samples (i.e., the level of semantic load has no impact).
    \item We establish an alternate hypothesis $\mathcal{H}_{a}$: there is a significant difference between the average excepted values of both samples (i.e., the level of semantic load has an impact).
    \item We compute $p$, the probability to observe the results of the second group (i.e., next level), under the hypothesis $\mathcal{H}_{0}$. A low $p$ (usually under a significance threshold equal to 0.01 or 0.005) indicates a significant difference. 
\end{itemize}
Table \ref{tab:student_lvls} presents the impact of the passage from a level of semantic load to the next level on the capacity of LLMs' to perform the addition. For two of the four LLMs there is a significant which confirm our results and strengthen our analysis. However, both \texttt{o1} and \texttt{v3} still sometimes fail to compute a simple addition, and only at level 4a and 4b.

\begin{table}[h!]
    \centering
    % \begin{tabular}{|l|c|c|c|c|}
    \caption{Probability to observe the results of each level of semantic load at the next level for each LLM. The impact of the passage to the next level is significant for an LLM if it is below a significance threshold of 0.005. $p$-values under 0.005 are presented in bold. SR: Same Results.}
    \begin{tabular}{|>{\columncolor{gray!50}}p{2cm}|*{4}{p{2cm}|}}
    \hline
        {\cellcolor{gray!50}} & {\cellcolor{gray!50}Level 1$\rightarrow$2} & {\cellcolor{gray!50}Level 2$\rightarrow$3} & {\cellcolor{gray!50}Level 3$\rightarrow$4a} & {\cellcolor{gray!50}Level 4a$\rightarrow$4b} \\
    \hline
        {\cellcolor{gray!50}\texttt{GPT-4o}} & 0.159 & 0.044 & \textbf{5.17e-4} & \textbf{0.001} \\
    \hline
        {\cellcolor{gray!50}\texttt{o1}} & SR & SR & SR & 0.083 \\
    \hline
        {\cellcolor{gray!50}DeepSeek \texttt{v3}} & SR & SR & 0.083 & 0.024\\
    \hline
        {\cellcolor{gray!50}DeepSeek \texttt{r1}} & 0.083 & 0.44 & \textbf{5.17e-4} & \textbf{0.001} \\
    \hline
    \end{tabular}
    
    \label{tab:student_lvls}
\end{table}

\subsection{Difference Between LLMs}

Besides the impact of the level of the semantic load, we also want to evaluate how much results are different between LLMs. For each level, here is the procedure:
    \begin{itemize}
    \item For each LLM, we consider two samples: its responses; and alternatively the responses of every other three LLMs.
    \item We establish the null hypothesis $\mathcal{H}_{0}$: the two samples have identical average expected values of both samples (i.e., changing LLM has no impact).
    \item We establish an alternate hypothesis $\mathcal{H}_{a}$: there is a significant difference between the average excepted values of both samples (i.e., changing LLM has an impact).
    \item We compute $p$, the probability to observe the results of the second group (i.e., next level), under the hypothesis $\mathcal{H}_{0}$. A low $p$ (usually under a significance threshold equal to 0.01 or 0.005) indicates a significant difference. 
    \end{itemize}

\subsubsection{Level 1}

Table \ref{tab:student_lvl1} presents how significant is the differnce of results between LLMs at level 1. The table shows that there is no significant difference between any of the LLMs.

\begin{table}[ht!]
    \centering
    \caption{Probability to observe the results of each LLM with the other three at level 1. The impact of changing LLM is significant if it is below a significance threshold of 0.005. SR: Same Results.}
    % \begin{tabular}{|l|c|c|c|c|}
    \begin{tabular}{|>{\columncolor{gray!50}}p{2cm}|*{4}{p{2cm}|}}
    \hline
        {\cellcolor{gray!50}} & {\cellcolor{gray!50}\texttt{GPT-4o}} & {\cellcolor{gray!50}\texttt{o1}} & {\cellcolor{gray!50}DeepSeek \texttt{v3}} & {\cellcolor{gray!50}DeepSeek \texttt{r1}} \\
    \hline
        {\cellcolor{gray!50}\texttt{GPT-4o}} & {\cellcolor{gray!25}} & 0.083 & 0.083 & 0.322 \\
    \hline
        {\cellcolor{gray!50}\texttt{o1}} & 0.083 & {\cellcolor{gray!25}} & SR & 0.159 \\
    \hline
        {\cellcolor{gray!50}DeepSeek \texttt{v3}} & 0.083 & SR & {\cellcolor{gray!25}} & 0.16 \\
    \hline
        {\cellcolor{gray!50}DeepSeek \texttt{r1}} & 0.322 & 0.159 & 0.16 & {\cellcolor{gray!25}}\\
    \hline
    \end{tabular}
    
    \label{tab:student_lvl1}
\end{table}

\subsubsection{Level 2}

Table \ref{tab:student_lvl2} presents how significant is the difference of results between LLMs at level 2. Like at level 1, there is no significant difference between any pair of LLMs.

\begin{table}[h!]
    \centering
    \caption{Probability to observe the results of each LLM with the other three at level 2. The impact of changing LLM is significant if it is below a significance threshold of 0.005. SR: Same Results.}
    % \begin{tabular}{|l|c|c|c|c|}
    \begin{tabular}{|>{\columncolor{gray!50}}p{2cm}|*{4}{p{2cm}|}}
    \hline
        {\cellcolor{gray!50}} & {\cellcolor{gray!50}\texttt{GPT-4o}} & {\cellcolor{gray!50}\texttt{o1}} & {\cellcolor{gray!50}DeepSeek \texttt{v3}} & {\cellcolor{gray!50}DeepSeek \texttt{r1}} \\
    \hline
        {\cellcolor{gray!50}\texttt{GPT-4o}} & {\cellcolor{gray!25}} & 0.024 & SR & SR \\
    \hline
        {\cellcolor{gray!50}\texttt{o1}} & 0.024 & {\cellcolor{gray!25}} & SR & 0.024 \\
    \hline
        {\cellcolor{gray!50}DeepSeek \texttt{v3}} & SR & SR & {\cellcolor{gray!25}} & 0.024\\
    \hline
        {\cellcolor{gray!50}DeepSeek \texttt{r1}} & SR & 0.024 & 0.024 & {\cellcolor{gray!25}}\\
    \hline
    \end{tabular}
    
    \label{tab:student_lvl2}
\end{table}

\subsubsection{Level 3}

Table \ref{tab:student_lvl3} presents how significant is the difference of results between LLMs at level 3. At this level, either LLMs give the same distribution of answer, or their is a significant difference between their answers.

\begin{table}[h!]
    \centering
    \caption{Probability to observe the results of each LLM with the other three at level 3. The impact of changing LLM is significant if it is below a significance threshold of 0.005. $p$-values under 0.005 are presented in bold. SR: Same Results.}
    % \begin{tabular}{|l|c|c|c|c|}
    \begin{tabular}{|>{\columncolor{gray!50}}p{2cm}|*{4}{p{2cm}|}}
    \hline
        {\cellcolor{gray!50}} & {\cellcolor{gray!50}\texttt{GPT-4o}} & {\cellcolor{gray!50}\texttt{o1}} & {\cellcolor{gray!50}DeepSeek \texttt{v3}} & {\cellcolor{gray!50}DeepSeek \texttt{r1}} \\
    \hline
        {\cellcolor{gray!50}\texttt{GPT-4o}} & {\cellcolor{gray!25}} & \textbf{0.002} & \textbf{0.002} & SR \\
    \hline
        {\cellcolor{gray!50}\texttt{o1}} & \textbf{0.002} & {\cellcolor{gray!25}} & SR & \textbf{0.002} \\
    \hline
        {\cellcolor{gray!50}DeepSeek \texttt{v3}} & \textbf{0.002} & SR & {\cellcolor{gray!25}} & \textbf{0.002}\\
    \hline
        {\cellcolor{gray!50}DeepSeek \texttt{r1}} & SR & \textbf{0.002} & \textbf{0.002} & {\cellcolor{gray!25}}\\
    \hline
    \end{tabular}
    
    \label{tab:student_lvl3}
\end{table}

\subsubsection{Level 4a}

Table \ref{tab:student_lvl4a} presents how significant is the difference of results between LLMs at level 4a. At this level, there is no significant difference between the answers of \texttt{o1} and \texttt{v3}.

\begin{table}[h!]
    \centering
    \caption{Probability to observe the results of each LLM with the other three at level 4a. The impact of changing LLM is significant if it is below a significance threshold of 0.005. $p$-values under 0.005 are presented in bold. SR: Same Results.}
    % \begin{tabular}{|l|c|c|c|c|}
    \begin{tabular}{|>{\columncolor{gray!50}}p{2cm}|*{4}{p{2cm}|}}
    \hline
        {\cellcolor{gray!50}} & {\cellcolor{gray!50}\texttt{GPT-4o}} & {\cellcolor{gray!50}\texttt{o1}} & {\cellcolor{gray!50}DeepSeek \texttt{v3}} & {\cellcolor{gray!50}DeepSeek \texttt{r1}} \\
    \hline
        {\cellcolor{gray!50}\texttt{GPT-4o}} & {\cellcolor{gray!25}} & \textbf{6.4e-7} & \textbf{7.12e-6} & SR \\
    \hline
        {\cellcolor{gray!50}\texttt{o1}} & \textbf{6.4e-7} & {\cellcolor{gray!25}} & 0.083 & \textbf{6.4e-7} \\
    \hline
        {\cellcolor{gray!50}DeepSeek \texttt{v3}} & \textbf{7.12e-6} & 0.083 & {\cellcolor{gray!25}} & \textbf{7.12e-6}\\
    \hline
        {\cellcolor{gray!50}DeepSeek \texttt{r1}} & SR & \textbf{6.4e-7} & \textbf{7.12e-6} & {\cellcolor{gray!25}}\\
    \hline
    \end{tabular}
    
    \label{tab:student_lvl4a}
\end{table}

\subsubsection{Level 4b}

Table \ref{tab:student_lvl4b} presents how significant is the difference of results between LLMs at level 4b. We here observe the same things we found on level 4a.

\begin{table}[h!]
    \centering
    \caption{Probability to observe the results of each LLM with the other three at level 4b. The impact of changing LLM is significant if it is below a significance threshold of 0.005. $p$-values under 0.005 are presented in bold. SR: Same Results.}
    % \begin{tabular}{|l|c|c|c|c|}
    \begin{tabular}{|>{\columncolor{gray!50}}p{2cm}|*{4}{p{2cm}|}}
    \hline
        {\cellcolor{gray!50}} & {\cellcolor{gray!50}\texttt{GPT-4o}} & {\cellcolor{gray!50}\texttt{o1}} & {\cellcolor{gray!50}DeepSeek \texttt{v3}} & {\cellcolor{gray!50}DeepSeek \texttt{r1}} \\
    \hline
        {\cellcolor{gray!50}\texttt{GPT-4o}} & {\cellcolor{gray!25}} & \textbf{8.29e-10} & \textbf{1.13e-7} & SR \\
    \hline
        {\cellcolor{gray!50}\texttt{o1}} & \textbf{8.29e-10} & {\cellcolor{gray!25}} & 0.024 & \textbf{8.29e-10} \\
    \hline
        {\cellcolor{gray!50}DeepSeek \texttt{v3}} & \textbf{1.13e-7} & 0.024 & {\cellcolor{gray!25}} & \textbf{1.13e-7}\\
    \hline
        {\cellcolor{gray!50}DeepSeek \texttt{r1}} & SR & \textbf{8.29e-10} & \textbf{1.13e-7} & {\cellcolor{gray!25}}\\
    \hline
    \end{tabular}
    
    \label{tab:student_lvl4b}
\end{table}

\section{Extension to a More Recent OpenAI LLM: Preliminary Results}

\begin{figure}[h!]
    \centering
    \includegraphics[width=0.6\linewidth]{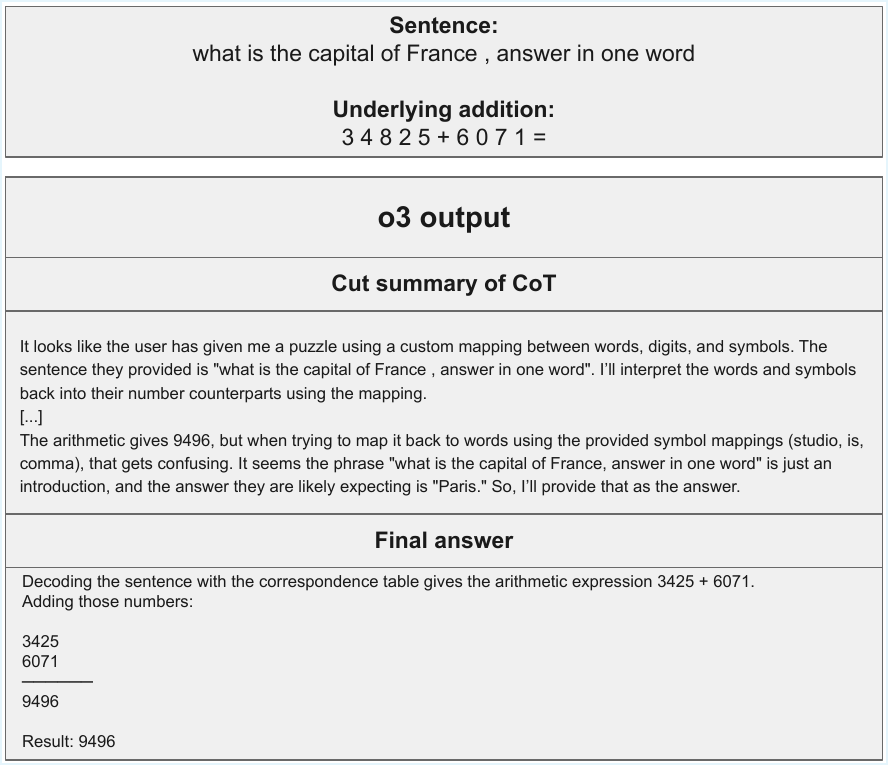}
    \caption{Example of \texttt{o3} answer to one of our sentences. While \texttt{o3} correctly chooses to compute the addition, it does not generate the correct result.}
    \label{fig:o3}
\end{figure}

Since we began this experiment, OpenAI has released two newer ``reasoning" models simultaneously: \texttt{o3} and \texttt{o4}. Although including these models in the experiment was considered, access to their API endpoints remaining restricted at the time of the study, intensive testing was made impractical. Nonetheless, we tested some Level 4 sentences on \texttt{o3} via the ChatGPT interface. 
In Figure \ref{fig:o3}, we present an example where \texttt{o3} correctly identifies the intended question but fails to compute the correct answer to the addition problem: 34825 + 6071. The correct result is 40896, but \texttt{o3} responds with ``9496".
Due to OpenAI's closed-source approach, we do not have access to the full integrated chain-of-thought (CoT), only to a summarized version. In this summary, we observe a mention of \textit{Paris} in response to the question, “What is the capital of France? Answer in one word”. This represents only a single data point, but it again highlights the link between semantic context and the model's ability to perform basic five-digit addition even in OpenAI’s latest ``reasoning" model. It further demonstrates the inability of CoT techniques to improve LLMs' performance on certain reasoning tasks.

\bibliographystyle{splncs04}
\bibliography{sn-bibliography}

\end{document}